# Deep Stochastic Radar Models

Tim A. Wheeler[1], Martin Holder[2], Hermann Winner[2], and Mykel J. Kochenderfer[1]

*Abstract*— Accurate simulation and validation of advanced driver assistance systems requires accurate sensor models. Modeling automotive radar is complicated by effects such as multipath reflections, interference, reflective surfaces, discrete cells, and attenuation. Detailed radar simulations based on physical principles exist but are computationally intractable for realistic automotive scenes. This paper describes a methodology for the construction of stochastic automotive radar models based on deep learning with adversarial loss connected to real-world data. The resulting model exhibits fundamental radar effects while remaining real-time capable.

## I. INTRODUCTION

Automotive radar is widely used within modern advanced driver assistance and vehicle safety systems. Current vehicles are equipped with up to six radar sensors enabling a 360° FOV in both near (up to 40 m) and far (up to 200 m) range [1], [2]. Unlike lidar or stereo video sensors, radar exploits the Doppler effect for obtaining relative velocity and is fairly robust to weather and lighting conditions. Radar sensors use radiated electromagnetic energy to measure the locations and velocities of objects in the sensor's field of view. The receiving antennae measure the returned power to infer distance and relative velocity based on how the modulated signal mutates during flight and reflection on targets. The raw radar measurement is then resolved into a three-dimensional grid of range, azimuth, and relative velocity.

Most commercial automotive radars provide an object list rather than the raw radar grid. The object list is typically the product of object clustering, ambiguity resolution, and target tracking, and is used to inform environmental models for driving assistance functions.

Radar exhibits several special characteristics, including multipath reflections, interference, ghost objects, ambiguities, clutter, and attenuation [3]. These effects are a result of high frequency electromagnetic waves propagating and interacting in complex scenes. The accurate simulation of such phenomena using physics-based models often requires computationally extensive ray-tracing calculations. Such approaches are currently not fast enough for practical use in simulation as they require highly detailed geometry (i.e. square millimeter accuracy) to obtain correct results, which require a massive amount of computational power. However, sensor models are needed to accurately represent radar in simulations of advanced driver assistance systems.

Existing radar models are typically black- or white-box. Black-box models seek to represent radar phenomena in a stochastic manner [4]. White-box models use ray-tracing for estimating electromagnetic path propagation and typically rely on object radar cross section values being given [5] or extract virtual scattering centers [6]. White-box models can exploit modern GPU technology for significantly faster computation [7]. However, such white-box models require detailed models of the environment to capture important radar-related effects such as multipath propagation and interference. Models of sufficient detail are often not available, and the extensive computation required makes real-time simulation infeasible. To the best knowledge of the authors, modern commercial software tools for virtual test driving use idealized sensor models, which do not reflect the sensor's actual performance.

This paper presents a new class of deep stochastic radar models for use in automotive simulations. These models allow for arbitrary roadway configurations and scene composition through the use of raster grid and object list inputs. These models produce power return fields that exhibit radar phenomena without explicit programming and run in real time. The proposed method can be seamlessly extended with additional object types and properties to capture additional behavior, and can be trained end-to-end. The methods were evaluated using prediction accuracy on withheld data, on reproducing the radar range equation for cube corner reflector measurements, and on predicted offroad clutter location and power distributions.

## II. MODEL SPECIFICATION

A sensor model $\mathcal{M}$ consists of a conditional probability distribution $p(Y \mid S)$ over the sensor observation $Y$ given a scene representation $S$. The scene $S$ describes the state of the environment relevant to the sensor, including the roadway geometry, scenery, road infrastructure, and other traffic participants. The observation $Y$ is a sensor reading, in this case a tensor over range and azimuth describing the received power by the radar over a discretized grid. A good sensor model produces observations that closely match those obtained in the real world.

Sensor models in simulation require a scene representation that is as close as possible to the data structures used by automotive simulation software. There are two primary ways in which scene information is conveyed: spatial rasters and object lists. Amorphous spatial information is often well represented by a spatial raster, in which the presence or absence of features can be indicated to a required spatial resolution [8]. Spatial rasters are useful for representing occupied space or marking the type of terrain at a particular grid point.

[1]Tim A. Wheeler and Mykel J. Kochenderfer are with Aeronautics and Astronautics, Stanford University, Stanford, CA 94305, USA {wheelert, mykel}@stanford.edu

[2]Martin Holder and Hermann Winner are with Institute of Automotive Engineering, Technische Universität Darmstadt, 64287 Darmstadt, Germany {holder, winner}@fzd.tu-darmstadt.de

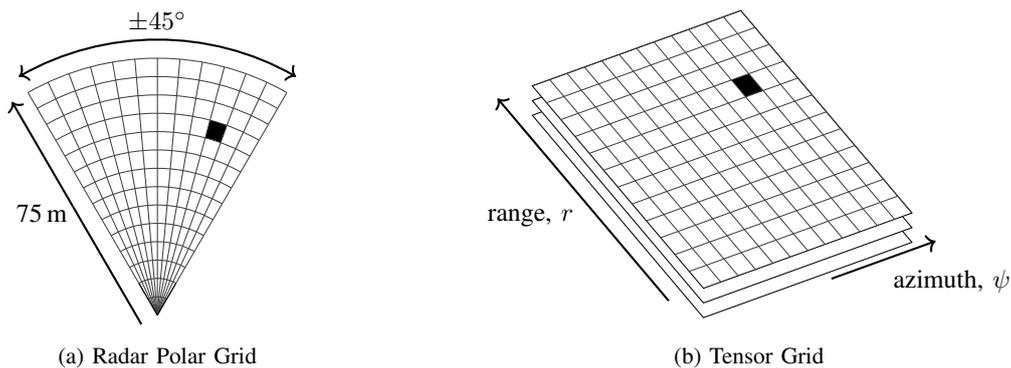

(a) Radar Polar Grid      (b) Tensor Grid

Fig. 1: The spatial raster input representation. The tensor is indexed using range and azimuth in the radar frame.

Object-specific information, especially dynamic entities like vehicles and pedestrians, are often represented using object lists. Each object has a list of properties that apply to the object as a whole, such as its pose information and material properties. Object lists allow for reasoning about specific entities, including a dynamic number of entities, and are easily extended with new properties. Foregoing an object list and using a raster instead requires rendering the properties to a discrete grid, leading to a loss of information for larger input sizes, and either requires many grid layers or a method for handling overlapping objects rendered to the same cell.

A sensor model should thus support a scene representation consisting of both components: a spatial raster $R$ and an object list $O$. These model inputs are both complex and multi-dimensional, as is required to capture the full space of driving scenes. Traditional parametric radar models do not have the modeling capacity to handle such complicated inputs and outputs [4], so we turn to deep learning.

## III. Deep Radar Model

Neural networks are universal function approximators that take the form of a computational graph [9]. Tensor-valued inputs are manipulated by a set of transformations as they traverse the directed acyclic network to produce tensors at the output nodes. These operations are parameterized, and can be efficiently trained with backpropagation via stochastic gradient descent [10].

Deep neural networks have recently gained widespread popularity, owing to their ability to automatically learn robust hierarchical features from complicated inputs [11], [12]. They have outperformed traditional state-of-the-art methods in fields as diverse as image classification [13] and natural language processing [14] and dominate the KITTI automotive benchmarks [15].

Neural networks are efficient to use once trained. Efficiency is critical for their application in simulation, in which

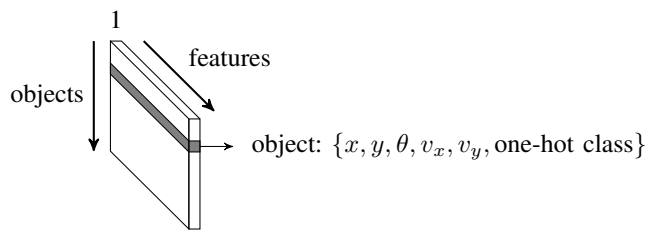

object: $\{x, y, \theta, v_x, v_y, \text{one-hot class}\}$

Fig. 2: The object list input allows for parameter sharing through $1 \times 1$ convolutions on a 3D tensor.

sensor readings must be generated for every vehicle in every simulation frame.

We employ deep neural networks as radar models to learn the conditional probability distribution $p(Y \mid R, O)$ end-to-end from data. The inputs to the network are the spatial raster $R$ and the object list $O$, and the output is the sensor reading $Y$.

### A. Inputs

The spatial raster $R$ is a 3D tensor with two primary dimensions of range and azimuth. The third dimension consists of layers that are dedicated to different types of information. This representation is quite similar to that of an RGB image whose pixel information is stored in two spatial dimensions and a color channel. Convolutions can be applied to such tensors in order to efficiently learn robust, hierarchical features [16].

Our spatial raster consists of a single layer indicating which cells contain grass and which cells contain roadway. More layers can be incorporated in the future as necessary. Such a representation allows for arbitrary roadway geometry that would be difficult to represent with an object list. Furthermore, the spatial raster is indexed in polar coordinates, mimicking the radar power grid, depicted in Fig. 1.

The object list $O$ is also represented as a 3D tensor (see Fig. 2). As with the spatial raster, we include a preprocessing head for learning abstract features before concatenation with the spatial raster.

The first processing layers for the object list are object-independent and designed for parameter sharing. This inde-pendence was accomplished using 2D convolutions, which also reduce the number of parameters and increase robustness [16]. The tensor has shape $n_{\text{objects}} \times 1 \times n_{\text{features}}$, where the first two dimensions correspond to traditional tensor length and width, and the features dimension corresponds to the number of channels. One-by-one convolutions are applied to the tensor, thereby sharing operations across all objects in the list.

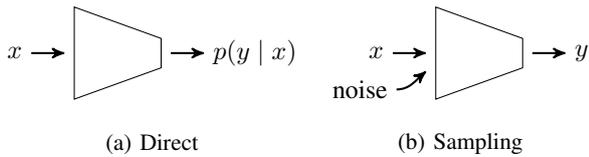

Fig. 3: The two typical methods for representing conditional probability distributions are by either directly parameterizing a probability distribution (a) or by producing samples using noise as an auxiliary input (b).

The object list representation is general and allows for the inclusion of arbitrary features. We include the object 2D pose $\{x, y, \theta\}$, speed, and a one-hot class encoding over objects classes. An additional class entry exists for when the object row is not used, which occurs when the list has more capacity than there are objects in a scene. When an object row is not used, all entries but the one-hot class entry are zeroed. Object geometry is not specified as it is given by the object class.

### B. Architectures

There are two primary methods[1] for representing a probability distribution with a neural network: by direct parameterization and by producing samples using noise as an auxiliary input. These approaches are referred to as the *direct* and *sampling* approaches, respectively. They are visualized in Fig. 3.

We provide two baseline radar models that directly parameterize a probability distribution: a normal distribution and a Gaussian mixture model. The multivariate normal distribution is often used in machine learning due to its well-behaved mathematics.

Unfortunately, the normal distribution is unimodal. Furthermore, the parameters for the normal distribution grow quadratically with the dimensionality of the target variable. One often sacrifices correlations between components and uses diagonal covariance matrices [18]. Here, the output of the neural network is a tensor grid with two layers; one for the mean and one for the diagonal log variances.

The second baseline radar model is a deep Gaussian mixture model (GMM) [19]:

$$p(x) = \sum_{i=1}^{n} w_i^\top \mathcal{N}(x \mid \mu_i, \sigma_i^2), \quad (1)$$

---

[1]Stochastic neural networks also exist [17], which introduce random variables directly within the network through stochastic activation functions or edge weights. These approaches are less efficient, as they require many samples both during training and at runtime.

where $w_i$, $\mu_i$, and $\sigma_i$ are the weight, mean, and standard deviation for the $i$th mixture component. The network outputs $3n$ layers to produce the weights, means, and log variances for each grid point. The component weights are squared and normalized to produce probability values, and the log variances are obtained through a ReLU [20] plus a small offset. GMMs have been successfully applied in fields such as speech synthesis [21] and generative models for image labels [18].

An important challenge with stochastic radar models is that the sensor outputs are multi-modal and spatially correlated. Regression approaches will average out the possible solutions, resulting in blurry predictions that do not resemble reality. A recently developed solution is the conditional variational autoencoder (VAE) [22], [23], which allows for learning one-to-many probability distributions without explicitly specifying the output distribution.

A VAE with input $X$ and output $Y$ seeks to model $p(Y \mid X)$. The model includes a latent random variable $z \sim \mathcal{N}(0, I)$ such that:

$$p(Y \mid X) = \mathcal{N}(Y \mid f(z, X), \sigma^2 I), \quad (2)$$

where $f$ is the deterministic neural network whose parameters are learned from data.

A function $Q(z \mid X, Y)$ is used to obtain a distribution over $z$ values that are likely to produce $X$. One minimizes the Kullback-Leibler divergence between $Q(z \mid X, Y)$ and $p(z \mid X, Y)$ to force it towards $\mathcal{N}(0, I)$:

$$\begin{aligned}\mathcal{D}[Q(z \mid X, Y) \, \| \, p(z \mid X, Y)] = \\ \mathbb{E}_{z \sim Q(\cdot \mid X, Y)}[\log Q(z \mid X, Y) - \log p(z \mid X, Y)]. \end{aligned} \quad (3)$$

Equation (3) is manipulated to produce [24]:

$$\begin{aligned}\log P(Y \mid X) - \mathcal{D}[Q(z \mid X, Y) \, \| \, p(z \mid Y, X)] = \\ \mathbb{E}_{z \sim Q(\cdot \mid X, Y)}[\log p(Y \mid z, X)] - \\ \mathcal{D}[Q(z \mid X, Y) \, \| \, p(z, X)]. \end{aligned} \quad (4)$$

The left hand side describes the optimization objective of $\log p(Y \mid X)$ with an error term for $Q$. The right hand side can be implemented as the loss function for stochastic gradient descent.

### C. Implementation Details

The overall architecture is an encoder-decoder pipeline shown in Fig. 4. The encoder takes the raster and object list and produces a latent feature representation $x$. The decoder takes the feature representation and a randomly generated noise value and produces the predicted sensor measurement. Source code for model training is released in the link at the end of the paper.

The encoder is composed of two heads, one for the spatial raster and one for the object list. These heads are joined and produce the latent feature representation. Both heads consist entirely of convolutional layers. The outputs are flattened, concatenated, and then processed using fully connected layers with ReLU activations [25].

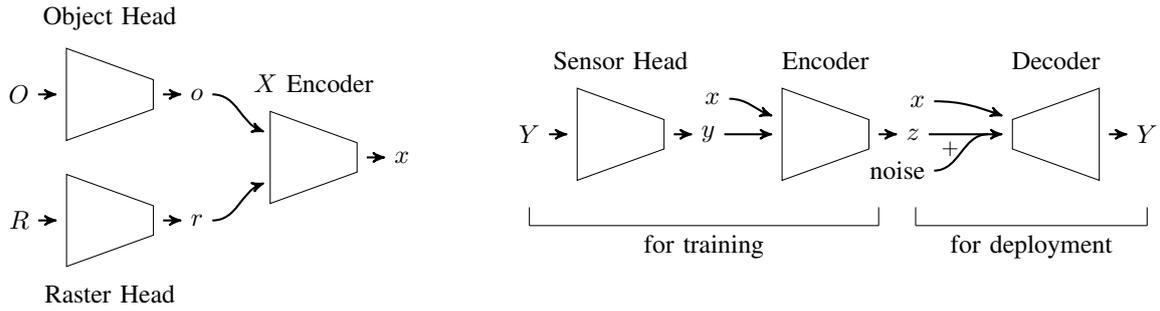

Fig. 4: Model architecture for the conditional variational autoencoder (VAE). The object list $O$ and raster grid $R$ each have their own processing heads that produce a flat vector. These inputs are concatenated and processed through a series of fully connected layers to produce the conditional input $x$. The VAE has both encoder and decoder stages that each take $x$ as an input. During training the latent variable $z$ depends on the sensor reading $Y$. When deployed, $z$ is sampled from a unit multivariate Gaussian.

The decoder generates a radar polar grid of power values using the encoded features and random noise. Random noise sampled from $\mathcal{N}(0, \sigma^2 I)$ is included as an input as per the VAE reparameterization trick with scalar hyperparameter $\sigma^2$ [26]. The noise and the latent features are concatenated and processed using fully connected layers with ReLU activations. The output is reshaped and followed by a series of deconvolutional layers [27] that increase dimensionality as convolutions with fractional stride. The repeated deconvolutions produce the required output dimensions of $64 \times 64 \times 1$.

### D. Loss Function

Neural networks are trained with stochastic gradient descent to minimize a scalar loss function [10]. The VAE loss,

$$\mathcal{L}_{\text{vae}} = \log p(Y \mid z, X, Y) - \mathcal{D}[Q(z \mid X, Y) \parallel p(z)], \quad (5)$$

includes a log-likelihood term for the prediction and a divergence term for the latent $z$ distribution. The VAE loss favors models that capture the overall structure of the radar signal with regard to its context, but tends to average together multiple modes and clutter objects in predictions.

An adversarial loss [28], however, trains a discriminator network $D(Y)$ to discriminate between real and generated data. The radar model can be trained to fool the discriminator according to a binary cross entropy loss, $\mathcal{L}_{\text{adv}}$. The adversarial loss forces the model to break away from smooth averages and commit to particular modes from its distribution.

The overall loss function for the deep radar model is:

$$\mathcal{L} = \alpha \mathcal{L}_{\text{vae}} + (1 - \alpha) \mathcal{L}_{\text{adv}} \quad (6)$$

where $\alpha \in [0, 1]$ adjusts the weights between the losses.

## IV. DATA COLLECTION

Data was collected using an automotive radar. Only the near scan range (up to 75 m) was considered. Experiments were run with the radar mounted on a moving vehicle.

An extensive set of recordings was taken on the German August-Euler airfield in sunny conditions. The runway provides a road-like corridor flanked by knee-high grass, which produces radar clutter at its boundary. Targets include a VW Golf, a cube corner reflector, a sedan-sized bag containing packing pellets, and a sedan-sized metal frame covered in black tarp, shown in Fig. 5. These targets possess very different radar signatures and require that the models learn to distinguish between them.

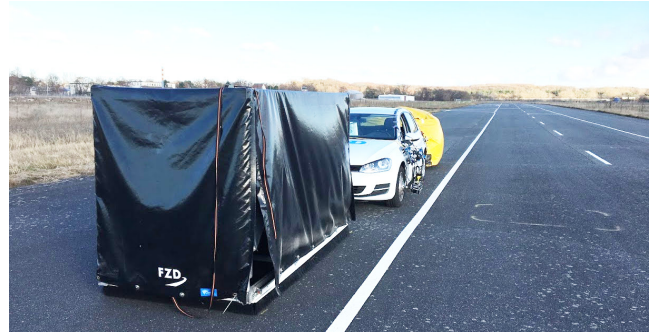

Fig. 5: Targets on the August-Euler airfield near Darmstadt, Germany. The yellow object is filled with packing pellets and the black object contains a metal tube frame. The VW Golf has attachments for measuring tire dynamics.

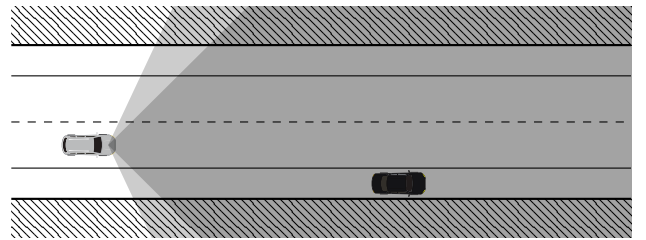

Fig. 6: Data collection with vehicle on runway using a radar with a 90° FOV and lidar with a 130° FOV.

Ground truth states were obtained from Lidar data fused from three Ibeo LUX laserscanners, as shown in Fig. 6. Radar power return points were exported for every frame. These points represent raw returns before ambiguity resolution, clustering, or object tracking is applied. Sensor readings for training and testing were constructed by rendering the radar points to the radar polar grid (Fig. 1) using their power in decibels. The terrain input was rendered to the radar polar grid and object lists were constructed from the derived ground truth. Every sample was extracted from over an hour of driving data at a 12.5 Hz sample rate, resulting in 47 359 frames.

TABLE I: Performance for each model on withheld data.

| model | expected root-mean square error |
|---|---|
| VAE (vae) | 23.23 |
| VAE (adv) | 25.12 |
| VAE (vae+adv) | 21.95 |
| Normal | 42.20 |
| GMM | 76.90 |

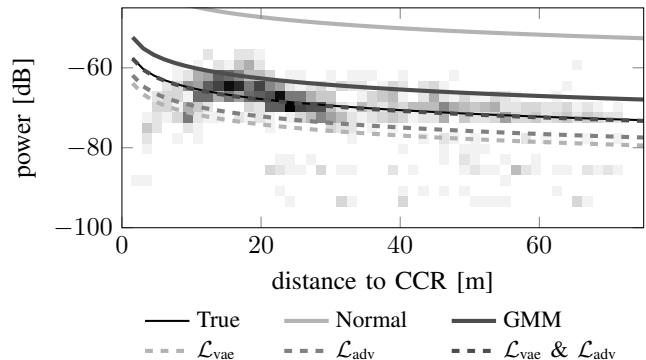

Fig. 7: Radar distance vs. range equations fit to withheld cube corner reflector data.

Real sensors exhibit noise and inaccuracies. Learning-based approaches are able to handle these inaccuracies directly and learn to exhibit the same characteristics. The sensor model should exhibit the same inaccuracies as the real sensors used to train it. Noise in the Lidar used to derive ground truth states, while minimal, was further mitigated by manually matching the Lidar point clouds with the radar measurements.

## V. EVALUATION

The conditional VAE radar models were evaluated on the collected dataset and compared against the baseline normal and Gaussian mixture models. All models were trained in Keras [29] with a Tensorflow [30] backend. We used the ADADELTA [31] stochastic gradient descent solver with mini-batches of size 16. Models were trained on a CUDA enabled GPU on a frame-wise random subset of 90% of the data and no model took longer than 12 h to train. Validation results are on the withheld data.

Three VAE models were trained, VAE (vae) based exclusively on the variational autoencoder loss, VAE (adv) based exclusively on the adversarial loss, and VAE (vae+adv) using a mixture with $\alpha = 0.99$. Here, $\alpha$ is close to one to account for the difference in scale between the two losses.

Qualitative results are shown in Fig. 8, depicting several withheld ground truth sample states, the true sensor reading, and the corresponding model predictions. The smoothing effect of the VAE loss is apparent, as is the uncorrelated nature of the two direct models. The models trained with adversarial loss exhibit multi-modality and concentrated power returns.

### A. Expected Root Mean Square Error

Generalization is evaluated by how well a model predicts withheld data. The expected root-mean square error of the sensor output for each model was computed by averaging the square deviation between the test set and one sample for each test input.

The results given in Table I show that the VAE models significantly outperform the normal and GMM methods. The mixed-loss VAE model had the lowest error. The Gaussian mixture model had higher error than the normal distribution, potentially due to overfitting.

### B. The Radar Range Equation

The radar range equation states that the power return for a radar is inversely proportional to the quartic distance to the target, $P_r \propto 1/r^4$ [3]. Cube corner reflectors (CCRs) were used as clear, standard targets for providing data to which power curves can be fit.

Power curves of the form $P_0/r^4$ were fit to minimize the square power, in decibels, error for each model's predicted CCR power returns on withheld data. These curves are shown in Fig. 7 overlaid on a 2D histogram of the true CCR returns. The results show a peak return at around 18 m, reflecting that the radar range equation is an approximation to more complicated phenomena and that the vehicle always drive past the CCR with different lateral offsets rather than stopping directly before it. Nevertheless, a radar range curve provides a reasonable fit to the data.

The mixed-loss VAE has nearly a perfect fit with the radar range equation fit to the ground truth. The pure VAE severely under-predicts the CCR power, owing to the fact that the VAE loss tends to smooth out predictions. The adversarial VAE also under-predicts the CCR power, but less severely. The adversarial network only receives the sensor reading as an input, and thus does not require the target to necessarily be in the correct location or have the correct radar return, so long as it results in an otherwise valid-looking reading. The two direct approaches over-predict the CCR return.

The results presented here only describe the phenomena in a qualitative manner. In future work, an experiment should be conducted in which one drives straight at a CCR.

### C. Clutter Metrics and Clutter Analysis

The radar power field often contains a significant number of returns caused by spurious reflections in the terrain. These clutter points are plainly visible in our data, occurring primarily at the grassy edge of the runway. Clutter is a real concern that should be mimicked by the model so that vehicles tested in simulation are affected as closely as they would be in the real world.

We evaluate the ability of each model to predict clutter using histograms over distance to clutter distance, heading to clutter, and power. A qualitative measure is obtained using the Kullback-Leibler divergence between the piecewise uniform distributions predicted by each model and the real-world data. A similar approach for radar model clutter analysis has been used by Bühren and Yang [32].

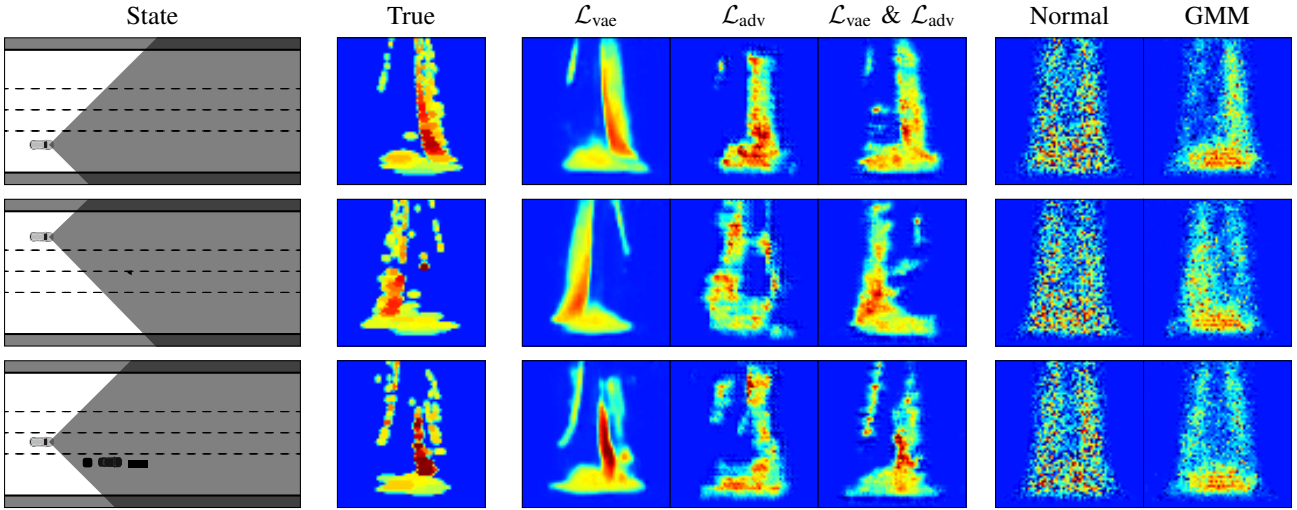

Fig. 8: Scene configurations with true radar measurements from withheld data and corresponding radar power field predictions. All heatmaps are tensor grid representations (Fig. 1) with power given in decibels.

TABLE II: Kullback-Leibler divergences for clutter.

| model | distance | angle | power |
| --- | --- | --- | --- |
| VAE (vae) | 0.027 | 0.219 | 0.141 |
| VAE (adv) | 0.055 | 0.243 | 0.368 |
| VAE (vae+adv) | 0.014 | 0.078 | 0.161 |
| Normal | 0.063 | 0.109 | 0.333 |
| GMM | 0.040 | 0.100 | 0.316 |

The clutter prediction histograms are given in Fig. 9. All models match fairly well in clutter distance and angle, indicating a high overall positional accuracy and suggesting that all models have learned to predict clutter values at the appropriate roadside locations. The associated KL-divergence values are listed in Table II, with the mixed-loss VAE performing best.

Clutter predicted power does not exhibit as good of a fit. The normal, GMM, and pure VAE models are trained to only minimize the regression loss, smoothing out the localized high power values. This is also exhibited by the dip in power return beneath −70 dB for true measurements in contrast with the high likelihood assigned by the models. Low power is over-predicted because the power return does not always exactly overlap with the true cluster position and the localized cluster power return is often smoothed over.

The adversarial VAE model has far more high power returns, whereas the mixed-loss VAE lies between the two VAE models, as expected. While the pure VAE has the lowest divergence in power, the gap between the second best, mixed-loss VAE model and the third best GMM model is significantly larger.

## VI. CONCLUSION

This paper developed deep radar sensor models for use in automotive simulation. An architecture was developed that closely mimics data structures used in automotive simulations and can be readily applied to arbitrary roadway topologies and scene configurations. Qualitative and quantitative evaluations against real-world data show that using conditional variational autoencoders trained with both autoencoder loss and adversarial loss produce the best predictions. The VAE is able to produce inter-correlated predictions, and the use of an adversarial discriminator prevents prediction averaging, resulting in more realistic predictions. Models were shown to exhibit power loss following the radar range equation and reproduce roadside clutter.

Future work will investigate additional radar sensor characteristics, including measurement accuracy, detection rate, interference, and separation capability. Model sensitivity to radar make and configuration should also be investigated.

A distinguishing feature of radar is its ability to measure relative velocities via the Doppler effect. Relative velocity should be incorporated into future versions of the deep radar model, perhaps by producing a three-dimensional output tensor with dimensions of range, azimuth, and relative velocity. The high dimensionality raises additional training concerns.

Perhaps the greatest limitation of a learning-based model is generalization to more complicated environments and the vast amounts of data needed to train such models. One potential solution is to develop and certify white-box models that may be too slow to run in real time but can be used to generate a large amount of data for training and validating a real-time deep stochastic radar model. These approaches should be tested against formal model requirements for specific use cases, such as in simulations for adaptive cruise control or autonomous driving, and the ability of the model to generalize should be quantified. All software is publicly available at `github.com/tawheeler/2017_iv_deep_radar`.

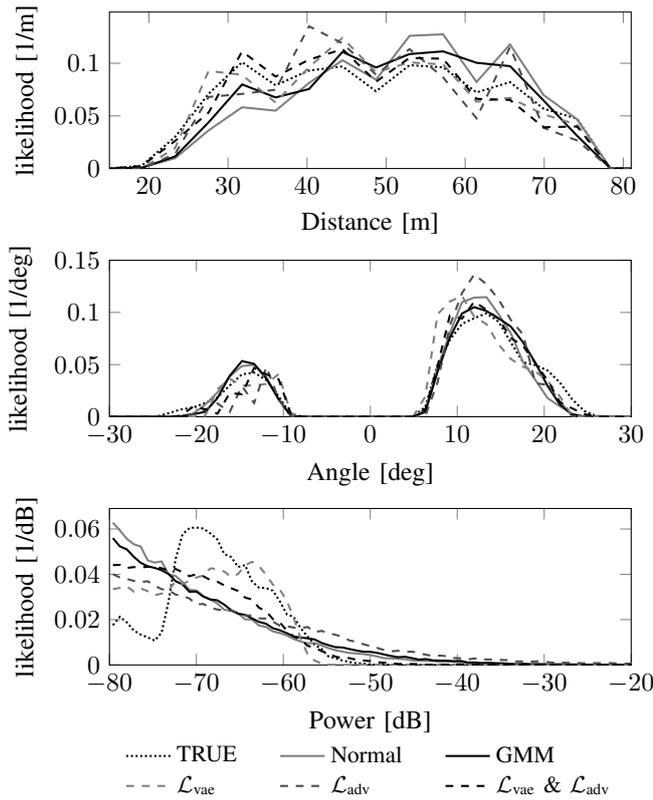

Fig. 9: Clutter prediction capability in distance, angle, and power. Only offroad reflections are included.

ACKNOWLEDGMENT

This work was partially funded by Robert Bosch LLC. The authors would like to thank Tobias Homolla for his support with data collection and TU Darmstadt for providing the Euler Flugplatz as a testing ground. We gratefully acknowledge the helpful comments received from anonymous reviewers.

REFERENCES

[1] J. Dickmann, N. Appenrodt, J. Klappstein, H.-L. Bloecher, M. Muntzinger, A. Sailer, M. Hahn, and C. Brenk, "Making Bertha see even more: Radar contribution", *IEEE Access*, vol. 3, pp. 1233–1247, 2015.
[2] H. Winner, "Automotive RADAR", *Handbook of Driver Assistance Systems: Basic Information, Components and Systems for Active Safety and Comfort*, pp. 325–403, 2016.
[3] M. I. Skolnik, *Radar handbook*, 3. ed. New York: McGraw Hill, 2008.
[4] S. Bernsteiner, Z. Magosi, D. Lindvai-Soos, and A. Eichberger, "Radarsensormodell für den virtuellen Entwicklungsprozess", *ATZelektronik*, vol. 10, no. 2, pp. 72–79, 2015.
[5] D. Gubelli, O. A. Krasnov, and O. Yarovyi, "Ray-tracing simulator for radar signals propagation in radar networks", in *European Radar Conference (EuRAD)*, 2013.
[6] K. Schuler, D. Becker, and W. Wiesbeck, "Extraction of virtual scattering centers of vehicles by ray-tracing simulations", *IEEE Transactions on Antennas and Propagation*, vol. 56, no. 11, pp. 3543–3551, 2008.
[7] J. M. Barney, "GPU advancements reduce simulation times for 25 GHz automotive radar models", in *European Conference on Antennas and Propagation (EuCAP)*, Piscataway, NJ: IEEE, 2015.
[8] S. Thrun, "Robotic mapping: A survey", *Exploring artificial intelligence in the new millennium*, vol. 1, pp. 1–35, 2002.
[9] K. Hornik, "Approximation capabilities of multilayer feedforward networks", *Neural networks*, vol. 4, no. 2, pp. 251–257, 1991.
[10] D. E. Rumelhart, G. E. Hinton, and R. J. Williams, "Learning representations by back-propagating errors", *Cognitive modeling*, vol. 5, no. 3, p. 1, 1988.
[11] H. Lee, R. Grosse, R. Ranganath, and A. Y. Ng, "Convolutional deep belief networks for scalable unsupervised learning of hierarchical representations", in *International Conference on Machine Learning (ICML)*, 2009.
[12] A. Krizhevsky, I. Sutskever, and G. E. Hinton, "Imagenet classification with deep convolutional neural networks", in *Advances in Neural Information Processing Systems (NIPS)*, Curran Associates, Inc., 2012, pp. 1097–1105.
[13] K. He, X. Zhang, S. Ren, and J. Sun, "Deep residual learning for image recognition", *arXiv preprint arXiv:1512.03385*, 2015.
[14] G. E. Dahl, D. Yu, L. Deng, and A. Acero, "Context-dependent pre-trained deep neural networks for large-vocabulary speech recognition", *IEEE Transactions on Audio, Speech, and Language Processing*, vol. 20, no. 1, pp. 30–42, Jan. 2010.
[15] A. Geiger, P. Lenz, and R. Urtasun, "Are we ready for autonomous driving? the kitti vision benchmark suite", in *IEEE Computer Society Conference on Computer Vision and Pattern Recognition (CVPR)*, 2012.
[16] Y. LeCun, L. Bottou, Y. Bengio, and P. Haffner, "Gradient-based learning applied to document recognition", *Proceedings of the IEEE*, vol. 86, no. 11, pp. 2278–2324, 1998.
[17] C. Turchetti, *Stochastic models of neural networks.* Amsterdam: IOS Press; Tokyo: Ohmsha, 2004.
[18] Y. Pu, Z. Gan, R. Henao, X. Yuan, C. Li, A. Stevens, and L. Carin, "Variational autoencoder for deep learning of images, labels and captions", in *Advances in Neural Information Processing Systems (NIPS)*, Curran Associates, Inc., 2016, pp. 2352–2360.
[19] C. M. Bishop, "Mixture density networks", Aston University, Birmingham, U.K., Tech. Rep. NCRG/4288, 1994.
[20] V. Nair and G. E. Hinton, "Rectified linear units improve restricted Boltzmann machines", in *International Conference on Machine Learning (ICML)*, 2010.
[21] H. Zen and A. Senior, "Deep mixture density networks for acoustic modeling in statistical parametric speech synthesis", in *International Conference on Acoustics, Speech, and Signal Processing (ICASSP)*, 2014.
[22] K. Sohn, H. Lee, and X. Yan, "Learning structured output representation using deep conditional generative models", in *Advances in Neural Information Processing Systems (NIPS)*, 2015.
[23] J. Walker, C. Doersch, A. Gupta, and M. Hebert, "An uncertain future: Forecasting from static images using variational autoencoders", in *European Conference on Machine Learning (ECML)*, 2016.
[24] C. Doersch, "Tutorial on variational autoencoders", *arXiv preprint arXiv:1606.05908*, 2016.
[25] R. H. Hahnloser, R. Sarpeshkar, M. A. Mahowald, R. J. Douglas, and H. S. Seung, "Digital selection and analogue amplification coexist in a cortex-inspired silicon circuit", *Nature*, vol. 405, no. 6789, pp. 947–951, 2000.
[26] D. P. Kingma and M. Welling, "Auto-encoding variational Bayes", *arXiv preprint arXiv:1312.6114*, 2013.
[27] M. D. Zeiler, D. Krishnan, G. W. Taylor, and R. Fergus, "Deconvolutional networks", in *IEEE Computer Society Conference on Computer Vision and Pattern Recognition (CVPR)*, 2010.
[28] I. Goodfellow, J. Pouget-Abadie, M. Mirza, B. Xu, D. Warde-Farley, S. Ozair, A. Courville, and Y. Bengio, "Generative adversarial nets", in *Advances in Neural Information Processing Systems (NIPS)*, 2014.
[29] F. Chollet, *Keras*, https://github.com/fchollet/keras, 2015.
[30] M. Abadi, A. Agarwal, P. Barham, E. Brevdo, Z. Chen, C. Citro, G. S. Corrado, A. Davis, J. Dean, M. Devin, *et al.*, "Tensorflow: Large-scale machine learning on heterogeneous systems", *arXiv preprint arXiv:1603.04467*, 2015.
[31] M. D. Zeiler, "Adadelta: An adaptive learning rate method", *arXiv preprint arXiv:1212.5701*, 2012.
[32] M. Bühren and B. Yang, "Simulation of automotive radar target lists considering clutter and limited resolution", in *International Radar Symposium*, 2007.